\documentclass[10pt,twocolumn,letterpaper]{article}

\usepackage[pagenumbers]{cvpr}      
\usepackage{bbding}
\usepackage{color}
\usepackage{graphicx}
\usepackage{xcolor}
\usepackage{colortbl, booktabs}
\usepackage{multicol}
\usepackage{multirow}
\usepackage{comment}

\usepackage{marvosym}

\definecolor{basefirst}{HTML}{BEE5BE} 
\colorlet{first1}{basefirst!7}
\colorlet{first2}{basefirst!14}
\colorlet{first3}{basefirst!21}
\colorlet{first4}{basefirst!99}

\definecolor{basesecond}{HTML}{8EC5FF}  
\colorlet{second1}{basesecond!9}
\colorlet{second2}{basesecond!25}
\colorlet{second3}{basesecond!40}
\colorlet{second4}{basesecond!80}

\definecolor{basethird}{HTML}{C3B1E1}
\colorlet{third1}{basethird!15}
\colorlet{third2}{basethird!25}
\colorlet{third3}{basethird!40}
\colorlet{third4}{basethird!55}

\definecolor{basefourth}{HTML}{FFDAB9}
\colorlet{fourth1}{basefourth!15}
\colorlet{fourth2}{basefourth!25}
\colorlet{fourth3}{basefourth!40}
\colorlet{fourth4}{basefourth!55}
\colorlet{fourth5}{basefourth!70}
\colorlet{fourth6}{basefourth!85}
\colorlet{fourth7}{basefourth!100}

\definecolor{zero4}{HTML}{FFDAB9}

\definecolor{cvprblue}{rgb}{0.21,0.49,0.74}
\usepackage[pagebackref,breaklinks,colorlinks,allcolors=cvprblue]{hyperref}

\title{Think Visually, Reason Textually: Vision-Language Synergy in ARC}

\author{Beichen Zhang$^{1,2}$, Yuhang Zang$^{2\textsuperscript{\Letter}}$, Xiaoyi Dong$^{1,2}$, Yuhang Cao$^{2}$\\ Haodong Duan$^{2}$, Dahua Lin$^{1, 2}$, Jiaqi Wang$^{2, 3\textsuperscript{\Letter}}$\\
$^1$The Chinese University of Hong Kong \quad
$^2$Shanghai AI Laboratory \quad 
$^3$Shanghai Innovation Institute \\
{\tt\small\{zhangbeichen,zangyuhang\}@pjlab.org.cn}
{\textsuperscript{\Letter}\small{Corresponding Authors.}}\\
}

\begin{document}
\maketitle
\begin{abstract}

Abstract reasoning from minimal examples remains a core unsolved problem for frontier foundation models such as GPT-5 and Grok 4.
These models still fail to infer structured transformation rules from a handful of examples, which is a key hallmark of human intelligence.
The Abstraction and Reasoning Corpus for Artificial General Intelligence (ARC-AGI) provides a rigorous testbed for this capability, demanding conceptual rule induction and transfer to novel tasks.
Most existing methods treat ARC-AGI as a purely textual reasoning task, overlooking the fact that humans rely heavily on visual abstraction when solving such puzzles.
However, our pilot experiments reveal a paradox: naively rendering ARC-AGI grids as images degrades performance due to imprecise rule execution.
This leads to our central hypothesis that vision and language possess complementary strengths across distinct reasoning stages:
vision supports global pattern abstraction and verification, whereas language specializes in symbolic rule formulation and precise execution.
Building on this insight, we introduce two synergistic strategies:
(1) Vision-Language Synergy Reasoning (VLSR), which decomposes ARC-AGI into modality-aligned subtasks; and
(2) Modality-Switch Self-Correction (MSSC), which leverages vision to verify text-based reasoning for intrinsic error correction.
Extensive experiments demonstrate that our approach yields up to a 4.33\% improvement over text-only baselines across diverse flagship models and multiple ARC-AGI tasks.
Our findings suggest that unifying visual abstraction with linguistic reasoning is a crucial step toward achieving generalizable, human-like intelligence in future foundation models. Source code is released at \url{https://github.com/InternLM/ARC-VL}.
\end{abstract}

\section{Introduction}

\begin{figure}[t]
  \centering
  \includegraphics[width=\linewidth]{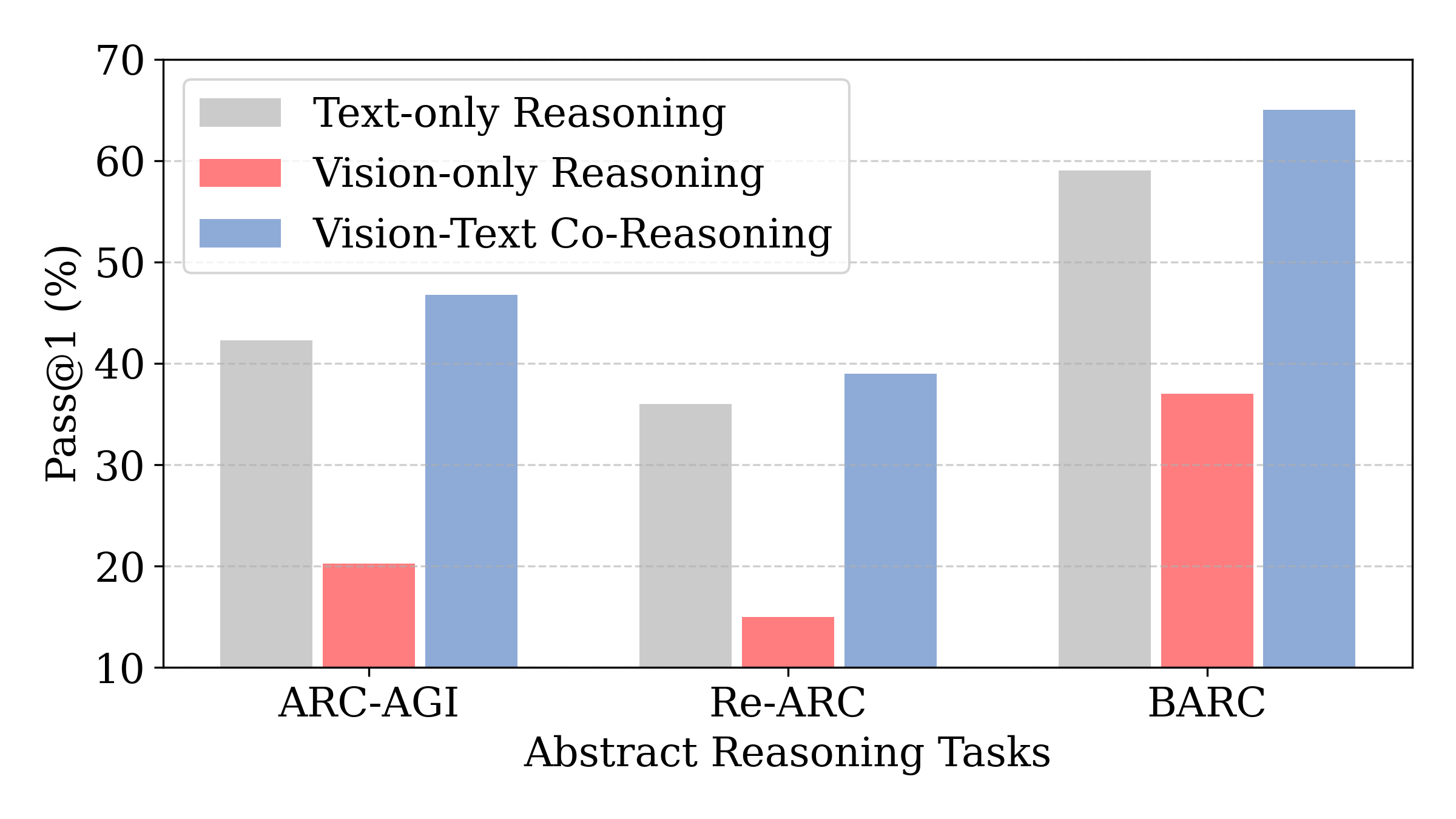}
  \vspace{-20pt}
   \caption{We propose vision-text co-reasoning in abstract reasoning tasks. It integrates the unique advantages of visual and textual thinking, thereby outperforming uni-modal reasoning. All methods use o4-mini as the base model.}
  \vspace{-16pt}
  \label{fig:bar}
\end{figure}
The Abstraction and Reasoning Corpus for Artificial General Intelligence (ARC-AGI) \cite{chollet2019measure} has emerged as the leading benchmark for evaluating machine intelligence beyond domain-specific skills.
Unlike traditional AI benchmarks that focus on narrow tasks such as natural language processing or image understanding, ARC-AGI measures the ability to \textit{learn how to learn}: a system must identify abstract transformation rules from minimal examples and apply them to entirely new scenarios.
This evaluation paradigm mirrors human IQ tests and represents a fundamental shift toward assessing artificial general intelligence.
Consequently, state-of-the-art models (e.g., GPT-5~\cite{openai2025gpt5} and Grok 4~\cite{xai2025grok4}) now highlight ARC-AGI performance as a key indicator of their reasoning capabilities.

\begin{figure*}[t]
  \centering
  \includegraphics[width=\linewidth]{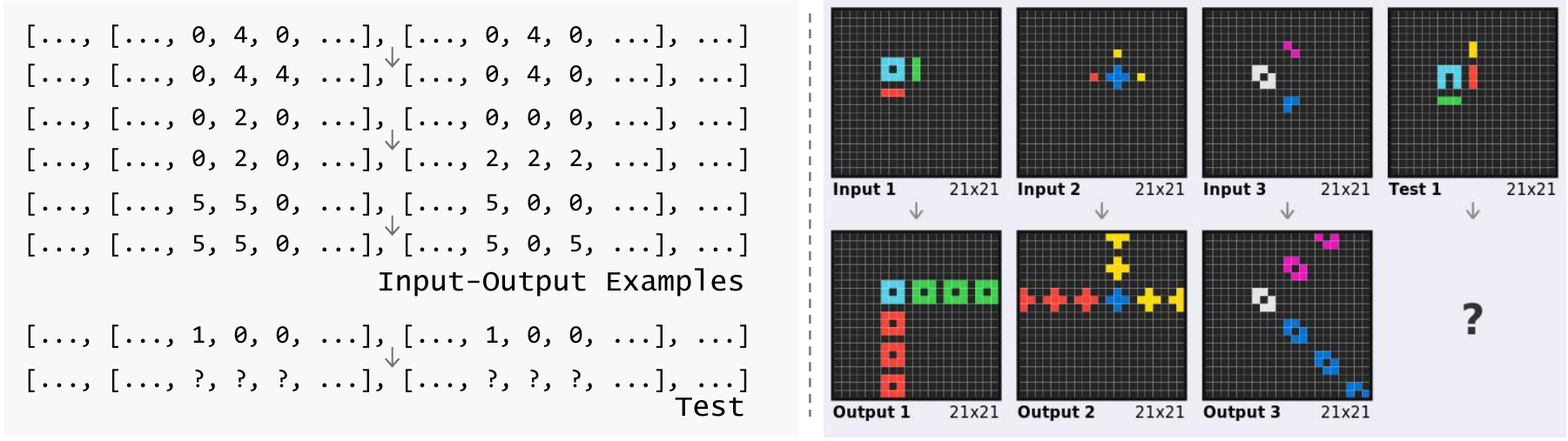}
  \vspace{-12pt}
   \caption{\textbf{Textual (left) vs. Visual (right) Thinking in the ARC-AGI Task.} Previous work treats ARC-AGI as a pure text task for training and reasoning, as text allows for a precise representation of each element. However, this approach loses the intuitiveness of visual thinking and 2D structural information. In contrast, we organically integrate visual thinking and textual thinking into the ARC-AGI reasoning process, using the complementary strengths of different modalities.
}
  \vspace{-12pt}
  \label{fig:text-vis}
\end{figure*}

Despite this importance, progress on ARC-AGI remains limited.
An intriguing observation is that \textbf{existing approaches treat ARC-AGI as a purely textual task}, representing the input-output matrix pairs as nested lists (e.g., [[0,1,2],[3,4,5]]) during both training and inference; see the left part of \cref{fig:text-vis}.
This design choice, while computationally convenient, fundamentally contradicts human problem-solving intuition.
As shown in the right part of \cref{fig:text-vis}, when humans approach ARC-AGI tasks, they naturally visualize the patterns: a color-coded 2D grid immediately reveals spatial relationships like symmetries, rotations, or shape transformations that are tedious to infer from textual coordinate descriptions.
We argue that this discrepancy between human visual intuition and machine text-centric processing represents a critical missed opportunity.

However, integrating visual information is not straightforward.
Our preliminary experiments reveal a counter-intuitive paradox: naively rendering ARC-AGI grids as images actually degrades performance compared to text-only baselines.
While image-based representations capture global 2D structure effectively, they struggle with precise element-wise operations.
For instance, when a $20\times20$ grid is presented as an image, models often fail to reliably identify or manipulate the value at a specific position such as $(5,7)$, conflating it with nearby cells.
This highlights a fundamental tension: vision excels at recognizing overall spatial patterns, whereas textual encodings naturally provide the discrete precision required for exact rule execution.

This observation leads to our core insight: \textbf{visual and textual modalities have fundamentally complementary strengths in different stages of abstract reasoning}.
To validate this hypothesis, we systematically decompose ARC-AGI into two sub-tasks: \textit{rule summarization} (extracting transformation patterns from examples) and \textit{rule application} (applying the extracted rule to new inputs).
We empirically evaluate each modality's performance on these sub-tasks (detailed in \cref{sec:modality_analysis}).
Our analysis of the OpenAI o4-mini model reveals striking differences: vision excels at rule summarization, providing a 3.0\% improvement through its holistic perception of 2D spatial structures, while text excels at rule application, with vision causing a dramatic 20.5\% performance drop due to imprecise element-wise manipulation.
These findings demonstrate that the question is not whether to use vision or text, but rather \textit{when and how} to strategically combine them.

Guided by our insights, we propose two synergistic strategies that strategically integrate visual and textual modalities throughout the abstract reasoning pipeline:

\textit{\textbf{V}ision-\textbf{L}anguage \textbf{S}ynergy \textbf{R}easoning (\textbf{VLSR})} matches each sub-task to its optimal modality.
During rule summarization, VLSR visualizes the example input-output matrix pairs as color-coded 2D grids, enabling the model to use holistic spatial perception and efficiently encode global transformation patterns (e.g., ``all shapes rotate 90 degrees clockwise'').
During rule application, VLSR switches to textual representation, allowing the model to perform precise element-wise manipulations guided by the extracted rule.
This modality-aware decomposition achieves improvements through two mechanisms: (1) divide-and-conquer task decomposition reduces individual sub-task complexity, and (2) strategic modality selection exploits each modality's inherent strengths.

\textit{\textbf{M}odality-\textbf{S}witch \textbf{S}elf-\textbf{C}orrection (\textbf{MSSC})} addresses a fundamental challenge in intrinsic self-correction: models struggle to identify errors when verifying their own reasoning in the same modality \cite{huang2023large,zhang2024understanding}.
MSSC breaks this limitation by employing different modalities for forward reasoning and backward verification. After generating a candidate output through text-based rule application, MSSC visualizes both the test input and predicted output as images, then uses the visual modality's strength in pattern consistency verification to check whether the predicted transformation matches the pattern shown in the example images.
If inconsistencies are detected, the model receives explicit feedback and performs another round of textual inference. This cross-modal verification enables effective intrinsic self-correction without any external information or ground truth.

Extensive experiments demonstrate that our combined approach achieves substantial improvements. 
When applied to flagship reasoning models such as Gemini-2.5-Pro and o4-mini, it delivers notable improvements of up to 7.25\%and 4.5\% accuracy respectively on the official ARC-AGI evaluation set. On average, our approach yields up to a 4.33\% improvement over text-only baselines across diverse models (GPT-4o, Gemini-2.5-Pro, o4-mini, Qwen3-VL) and multiple ARC-AGI benchmarks (ARC-AGI, BARC, Re-ARC).

Furthermore, our analysis reveals that text-only self-correction often fails or degrades performance, while MSSC provides consistent iterative improvements.

Our work makes the following contributions:

\textbf{1)} We provide the first systematic study of visual versus textual reasoning in ARC-AGI, identifying four key characteristics of their complementary strengths: holistic vs. independent processing, 2D structure preservation, encoding efficiency, and element-wise precision trade-offs, which is a novel perspective for the exploration of visual intelligence.

\textbf{2)} We introduce VLSR and MSSC, two training-free strategies that strategically combine visual and textual modalities to exploit their complementary strengths throughout the reasoning pipeline.

\textbf{3)} We demonstrate that our proposed vision-language synergy principle extends naturally to the training paradigm. By fine-tuning separate models for visual rule summarization and textual rule application, our approach achieves a 3.5\% improvement over text-only fine-tuning on the same training data, enabling small open-source models (Qwen3-8B) to surpass closed-source models like GPT-4o.
\section{Related Work}

\noindent \textbf{ARC-AGI Tasks.} ARC-AGI (Abstraction and Reasoning Corpus for Artificial General Intelligence) is a benchmark task designed to evaluate the generalization capability of AI systems. Its core objective is to measure whether an AI can abstract rules from a minimal number of examples and solve entirely new problems, rather than relying on large-scale datasets or pre-trained knowledge. Each task consists of several input-output matrix pairs. The model is required to infer rules from a small set of demonstrations and apply these rules to an unseen test input matrix. For humans, ARC-AGI tasks are not difficult, with an accuracy rate of over 97\%. However, they remain highly challenging for AI systems and have become one of the most demanding benchmarks currently used to assess whether an AI possesses artificial general intelligence.

\noindent \textbf{ARC-AGI Strategies.}  The Abstraction and Reasoning Corpus (ARC-AGI) task~\cite{chollet2019measure} has garnered widespread attention in the academic community since its proposal. Most strategies attempt to enhance ARC-AGI capabilities through training. A number of approaches~\cite{li2024combining,hodel2024addressing, moskvichev2023conceptarc,franzen2025product,park2023unraveling,xu2025neural,lee2024arcle,wang2025improving,moffitt2025arc,wang2024advancing,legris2024h,chen2025enigmata} generate a large number of ARC-AGI tasks by permuting and combining predefined transformation rules, and use this synthetic data to fine-tune large language models. Additionally, test-time training~\cite{sun2020test} is a prevalent strategy. These approaches~\cite{akyurek2024surprising, franzen2024llm, cole,pourcel2025self,zhut5,zuo2025ttrl} treat the input-output examples provided during testing not only as context to support the models' reasoning, but also as data for additional fine-tuning of model before it generates answers. Other strategies attempt to introduce additional hints during the inference process to enhance the model's performance on the ARC-AGI task. A common type of strategy~\cite{wang2023hypothesis, wind2020dsl, qiu2023phenomenal,singhal2025conceptsearch,macfarlane2024searching} leverages the characteristics of ARC-AGI tasks by predefining some of the potential transformation rules and then provides prompts and guidance to the model during the reasoning process. Another common strategy~\cite{ho2025arcmemo,suzgun2025dynamic, pourcel2025self, butt2024codeit,wang2025hierarchical,jolicoeur2025less} involves utilizing the model's memory: summarizing observations and insights from previous problems. This strategy consolidates previous attempts into referenceable concepts to guide the model in new reasoning tasks.
However, previous strategies have treated ARC-AGI as a pure text task for both training and reasoning, neglecting image representations that demonstrate the rules more intuitively.
Therefore, the core work of this paper is to integrate visual information into ARC-AGI and analyze the differences between visual thinking and textual thinking in abstract reasoning.

\noindent \textbf{Aiding Reasoning with Images.} Leveraging visual modalities to augment reasoning has emerged as a pivotal research direction. Visual Sketchpad~\cite{hu2024visual} enhances geometric problem-solving by rendering diagrams and introducing auxiliary constructions. ViLaSR~\cite{wu2025reinforcing} further proposes a ``Drawing to Reason in Space'' paradigm that permits the model to visualize its intermediate spatial hypotheses, achieving measurable improvements on spatial-reasoning benchmarks like maze solving. Owing to its immediacy and spatial expressiveness, visual information is becoming an indispensable component of reasoning pipelines.

\section{Method}

\begin{figure*}[t]
  \centering
  \includegraphics[width=\linewidth]{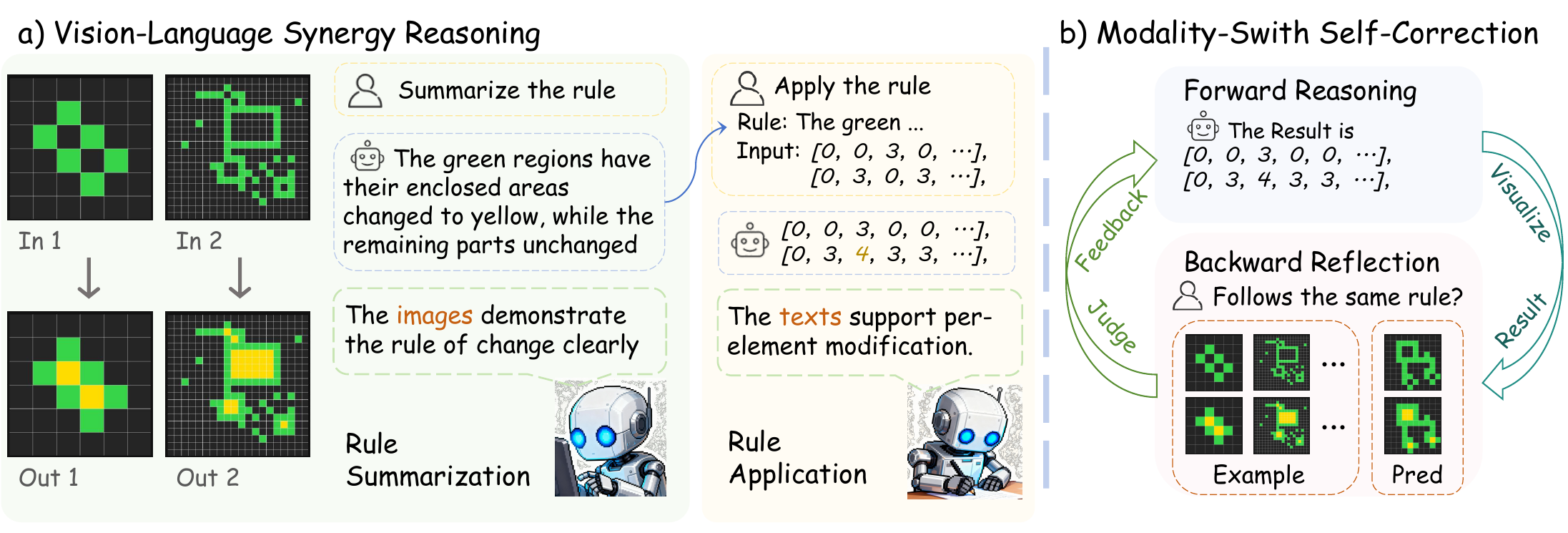}
  \vspace{-16pt}
   \caption{\textbf{Overview of our method.} \textbf{a) Vision-Language Synergy Reasoning} decomposes ARC-AGI into two subtasks: Rule-summarization and Rule-application. The former visualizes the provided example matrices as images, using global visual perception and 2D structure to summarize the rule. The latter requires element-wise processing, so rule-application is carried out in the textual modality. \textbf{b) Modality-Switch Self-Correction} visualizes the output matrix to judge rule consistency. The results are fed back to implement the self-correction strategy if necessary. As visual information is more informative in rule verification, the model can repeatedly refine its answers without relying on additional inputs.}
   \vspace{-12pt}
   \label{fig:overview}
\end{figure*}

The core of this work lies in integrating visual thinking into ARC-AGI tasks to use the \textbf{complementary strengths} of different modalities.
While existing approaches treat ARC-AGI as a pure text task, we argue that visual and textual modalities excel at different aspects of the reasoning process.
Visual thinking provides global perception and 2D structural understanding that text-centric reasoning lacks, while textual thinking enables precise element-wise processing and manipulation.

To this end, we propose two core methods for ARC-AGI (\cref{fig:overview}): \textbf{a)} Vision-Language Synergy Reasoning (VLSR) and \textbf{b)} Modality-Switch Self-Correction (MSSC).
VLSR summarizes rules from exemplar matrices in the visual modality and then applies the inferred rule to the test input in the text modality.
MSSC verifies the textual output by re-encoding it as an image to assess visual consistency with the examples; if inconsistent, it triggers another textual pass with targeted feedback.

We first formalize the ARC-AGI task and establish notation in \cref{sec:problem_setup}.
Before presenting our method design, we conduct a systematic empirical analysis in \cref{sec:modality_analysis} to understand the respective strengths of visual and textual modalities in ARC-AGI reasoning.
This provides the empirical foundation for our design choices.
Then we present the detailed design of VLSR in \cref{sec:vlsr} and MSSC in \cref{sec:mssc}.

\subsection{Problem Setup and Notation}\label{sec:problem_setup}

\noindent \textbf{Task Formulation.} Each ARC-AGI task is associated with a specific transformation rule $r$ that maps input matrices $m^{\text{input}}$ to output matrices $m^{\text{output}}$.
The model is provided with $K$ example pairs: $\{(m_1^{\text{input}}, m_1^{\text{output}}), (m_2^{\text{input}}, m_2^{\text{output}}), \ldots, (m_K^{\text{input}}, m_K^{\text{output}})\}$, where each matrix $m \in \mathbb{Z}^{H \times W}$ has $H, W \leq 30$ and cell values in $\{0, 1, \ldots, 9\}$.
The goal is to infer the underlying rule $r$ from these examples and apply it to a new test input $m_{\text{test}}^{\text{input}}$ to generate the corresponding output $m_{\text{test}}^{\text{output}}$.

\noindent \textbf{Notation.} We denote matrices in different modalities as follows:
\textbf{1)} $m$: the matrix, in any modality.
\textbf{2)} $t = \mathcal{T}(m)$: \textbf{textual representation} of the matrix (e.g., nested list: [[0,1,2],[3,4,5],[2,3,5]]).
\textbf{3)} $i = \mathcal{V}(m)$: \textbf{visual representation} of the matrix, where each cell value (0-9) is mapped to a distinct color in a grid layout.
Details of the visualization process are provided in \cref{app:visualization} in the supplementary materials.

The symbols $\mathcal{T}$ and $\mathcal{V}$ denote the text and visual transformation functions, respectively. Both transformations are invertible: $\mathcal{T}^{-1}(t) = m$ and $\mathcal{V}^{-1}(i) = m$, allowing seamless conversion between modalities.

\noindent \textbf{Previous Approaches.} Prior works process ARC-AGI tasks purely in the textual modality.
Besides, the transformation rule $r$ is not explicitly derived, either implicitly outputting it in the chain-of-thought, or skipping it directly to output the transformed matrix.

Formally, previous methods optimize a function $f$ to directly predict the output matrix in textual form:
\begin{equation}\label{eq:prev}
    t_{\text{pred}} = f(t_1^{\text{input}},t_1^{\text{output}}, \cdots, t_K^{\text{input}}, t_K^{\text{output}}, t_{\text{test}}^{\text{input}}).
\end{equation}

This has inherent \textbf{limitations}: (1) textual representation loses crucial 2D structural information, and (2) conflating rule summarization and application in a single step prevents the model from fully leveraging modality-specific strengths.

\subsection{Comparative Analysis of Vision and Text}
\label{sec:modality_analysis}

\begin{table}[!t]
  \caption{Quantitative experiments on using different modalities in rule summarization and application phase.}
  \vspace{-6pt}
  \centering
  \resizebox{0.49\textwidth}{!}{
  \normalsize
  \begin{tabular}{l|c|c|c|c|c}
    \toprule
    \multirow{2}{*}{Models} & \multirow{2}{*}{Baseline} & \multicolumn{2}{c|}{Rule-Sum.} & \multicolumn{2}{c}{Rule-App.} \\
    \cmidrule(lr){3-4} \cmidrule(lr){5-6} 
     &  & text & vision & text & vision \\
    \midrule
     GPT-4o & 8.25 & 10.5 & 13.5 & 13.5 &6.25\\
     Gemini-2.5 & 35.0 & 35.25 & 38.75 & 38.75 & 23.75 \\
     o4-mini & 42.25 & 42.5 & 45.5 & 45.5 & 25.0 \\
    \bottomrule
  \end{tabular}
  }
  \vspace{-12pt}
  \label{tab:analysis}
\end{table}

To inform our method design, we conduct a systematic empirical analysis of visual versus textual modalities in ARC-AGI reasoning.
We decompose the task into two subtasks: \textit{rule summarization} and \textit{rule application}, and evaluate each modality's performance on these subtasks separately across multiple state-of-the-art models.

\subsubsection{Quantitative Comparison}\label{sec:quant_analysis}

\noindent \textbf{Experimental Design.}
To isolate the effect of modality choice on each sub-task, we conduct experiments where we vary the modality used in either the rule summarization or rule application phase while keeping other factors constant.

\noindent \textit{Rule Summarization Phase.}
We input the provided example matrix pairs $\{(m_1^{\text{input}}, m_1^{\text{output}}), \ldots, (m_K^{\text{input}}, m_K^{\text{output}})\}$ into the same model using either textual representations $\{(t_i^{\text{input}}, t_i^{\text{output}})\}$ or visual representations $\{(i_i^{\text{input}},i_i^{\text{output}})\}$, and require the model to summarize the transformation rule $r$.
To fairly compare the quality of the extracted rules, we then uniformly apply both rules in the textual modality and compare the final accuracy on the test matrix.

\noindent \textit{Rule Application Phase.} For this phase, we adopt the same high-quality rule $r$ derived from the visual modality (as validated by the previous experiment).
The core comparison is whether it is better to represent both the example matrices and the test input matrix as images or as text when applying the rule to generate the output.

\noindent \textit{Baseline.} We also include the standard text-only approach commonly used in previous studies, which relies solely on the textual modality without explicit rule extraction, directly outputting the transformed test matrix.

\noindent \textbf{Results.} As shown in \cref{tab:analysis}, using the visual modality for rule summarization provides a clear advantage, yielding an average improvement of 3.2\% across models (e.g., 40.75\% vs. 37.25\% for Gemini-2.5).
However, this effect is reversed in the rule application phase.
When applying rules using visual representations instead of textual ones, performance drops dramatically by an average of 15.0\% (e.g., from 40.75\% to 23.75\% for Gemini-2.5).

Our results clearly demonstrate that \textbf{visual and textual modalities have complementary strengths}: vision excels at global pattern recognition needed for rule summarization, while text excels at precise element-wise manipulation needed for rule application.

\subsubsection{Qualitative Analysis: Understanding the ``Why''}
\label{sec:qual_analysis}

To understand the underlying reasons behind our quantitative findings, we conduct an in-depth qualitative analysis of model outputs across both modalities.
We identify four key characteristics that explain the performance differences:

\noindent \textbf{1) Visual thinking adopts holistic perception; textual thinking processes elements independently.}
Visual reasoning demonstrates a systematic bias toward encoding relational properties anchored on contiguous spatial structures, such as central blocks, checkerboard patterns, or connected components.
In contrast, textual reasoning relies more on type-level statistics (e.g., frequency counts) to identify patterns, treating each element more independently.
For rule summarization that requires identifying global spatial relationships, the holistic nature of visual thinking aligns better with task requirements and mirrors human cognitive strategies.

\noindent \textbf{2) Visual thinking preserves 2D structure; textual thinking may lose spatial information.}
In a matrix, two vertically adjacent elements in the same column are perceptually contiguous in a visual representation, yet may be separated by dozens of tokens in a textual representation (e.g., [[0,1,2],\textbf{[3]},4,5]] where elements in different rows are far apart in the token sequence).
Consequently, rules extracted through textual reasoning tend to lack 2D structural characteristics and perform poorly when tasks require capturing inter-row or diagonal regularities.
Moreover, when both input and output matrices are transposed, rules discovered via visual representation remain essentially invariant, while those derived from textual representation may be significantly affected by the change in token ordering.

\noindent \textbf{3) Visual representation is more efficient for large matrices.}
Textual representation requires separately encoding each matrix element along with delimiters (brackets, commas).
For large matrices (e.g., $30 \times 30$), textual representation may require thousands of tokens.
In contrast, visual representation encodes even complex matrices using a single image with only a few hundred vision tokens.
This enables faster reasoning without compromising or even enhancing the reasoning quality.
Recent work DeepSeek-OCR \cite{wei2025deepseekocrcontextsopticalcompression} also explores visual-based approaches for compressing complex document understanding.

\noindent \textbf{4) Visual reasoning lacks fine-grained element-wise precision.}
Images represent matrices as an integrated whole rather than encoding individual elements separately.
While this provides a global perspective advantageous for rule summarization, it becomes inadequate when element-wise processing is required during rule application.
We observe that when images represent large matrices, models may make errors in basic element localization and value identification (e.g., confusing the value at position $(5,7)$ with a nearby cell).
This limitation leads to performance degradation when visual thinking is adopted for the rule application, which demands precise per-element manipulation.

\noindent \textbf{Summary.} Our analysis reveals that visual and textual modalities have fundamentally different information processing characteristics.
Visual thinking provides global perception, preserves 2D spatial structure, and offers efficient representation, making it ideal for rule summarization.
Textual thinking enables precise element-wise access and manipulation, making it essential for rule application.
These complementary strengths motivate our method design in the following subsections.

\subsection{Vision-Language Synergy Reasoning}\label{sec:vlsr}

\noindent \textbf{Motivation.} We now translate the empirical insights from \cref{sec:modality_analysis} into our pipeline design.
Rather than forcing a single modality throughout the reasoning process, we introduce Vision-Language Synergy Reasoning that strategically switches modalities between subtasks.
Our key design principle is to \textit{route each sub-task to its optimal modality}: visual reasoning for rule summarization (exploiting global pattern recognition and 2D structural understanding) and textual reasoning for rule application (exploiting precise element-wise manipulation).
This modality-aware decomposition allows us to harness the full potential of LVLMs.

\noindent \textbf{Method Overview.} \cref{fig:overview} \textbf{(a)} presents our Vision-Language Synergy Reasoning (VLSR) pipeline, which strategically employs appropriate modalities at different stages. The complete pipeline consists of two phases:

\noindent \textit{\textbf{Phase 1}: Visual Rule Summarization.} We convert all example matrix pairs into \textbf{visual representations}.
Each matrix $m$ is visualized as an image $i = \mathcal{V}(m)$, where $\mathcal{V}$ is a visualization function that maps each cell value to a distinct color in a grid layout.
The LVLM then analyzes these visualized examples to derive an explicit transformation rule:
\begin{equation}
r_{\text{pred}} = f^{\text{vision}}_{\text{sum}}(i_1^{\text{input}}, i_1^{\text{output}}, i_2^{\text{input}}, i_2^{\text{output}}, \ldots, i_K^{\text{input}}, i_K^{\text{output}}),
\label{eq:rule_sum}
\end{equation}
where $f^{\text{vision}}_{\text{sum}}$ represents the LVLM operating in visual mode with a rule summarization prompt.
The derived rule $r_{\text{pred}}$ in \cref{eq:rule_sum} is expressed in natural language, describing the transformation pattern (e.g., ``rotate each connected component 90 degrees clockwise'').

\noindent \textit{\textbf{Phase 2}: Textual Rule Application.}
Given the summarized rule $r_{\text{pred}}$, we apply it to the test input using textual representations.
All matrices are converted to text format, and the model performs element-wise reasoning:
\begin{equation}
t_{\text{pred}} = f^{\text{text}}_{\text{app}}(r_{\text{pred}}, t_1^{\text{input}}, t_1^{\text{output}}, \ldots, t_K^{\text{input}}, t_K^{\text{output}}, t_{\text{test}}^{\text{input}}),
\label{eq:rule_app}
\end{equation}
where $f^{\text{text}}_{\text{app}}$ represents the same LVLM operating in text mode with a rule application prompt.
Note that $f^{\text{vision}}_{\text{sum}}$ and $f^{\text{text}}_{\text{app}}$ are the \textbf{same} base model; only the input modality and prompting strategy differ.

\noindent \textbf{Key Advantages.} Compared with \cref{eq:prev}, our approach offers two key benefits: (1) \textit{Task decomposition} reduces the complexity of individual subtasks through a divide-and-conquer strategy, and (2) \textit{Modality matching} ensures each sub-task uses the optimal modality: visual for global pattern recognition, textual for precise manipulation.
As we will demonstrate in \cref{sec:exp}, our strategy yields consistent improvements across different models and benchmarks.

\subsection{Modality-Switch Self-Correction}\label{sec:mssc}

\noindent \textbf{Motivation.} Self-correction is a cornerstone of human intelligence, yet it remains challenging for LLMs. The fundamental paradox is: if a model can identify and fix its own errors, why not generate the correct answer initially?
Existing works \cite{huang2023large,zhang2024understanding} have shown that intrinsic self-correction (without external feedback) is difficult because models struggle to distinguish correct from incorrect outputs when using the \textbf{same} reasoning modality.

We propose \textbf{M}odality-\textbf{S}witch \textbf{S}elf-\textbf{C}orrection \textbf{(MSSC)}, which achieves effective intrinsic self-correction by employing \textbf{different} modalities for reasoning and verification.
The key insight is that visual and textual modalities have complementary verification capabilities: while text excels at forward rule application, vision excels at pattern consistency verification.
By switching modalities, the model gains a fresh perspective that enables it to identify errors that are imperceptible in the original modality.

\noindent \textbf{Method Design.} \cref{fig:overview} \textbf{(b)} shows the pipeline.
After obtaining a candidate output $t_{\text{pred}}$ from textual rule application in \cref{eq:rule_app}, we perform the following iterative refinement:

\noindent \textit{Step 1: Visualization.} Convert the test input-output pair to visual form.
Since the predicted output $t_{\text{pred}}$ from \cref{eq:rule_app} is in textual format (nested list), we first parse it back to matrix form $m_{\text{pred}}$, then apply the visualization function $\mathcal{V}$:
\begin{equation}
    i_{\text{test}}^{\text{input}} = \mathcal{V}(t_{\text{test}}^{\text{input}}), \quad i_{\text{pred}} = \mathcal{V}(t_{\text{pred}}),
\end{equation}

\noindent \textit{Step 2: Visual Consistency Verification.} Present the visualized test pair with the examples to the LVLM as a critic:
\begin{equation}
    s_{\text{consistent}} = f^{\text{vision}}_{\text{critic}}(i_1^{\text{input}}, i_1^{\text{output}}, \ldots, i_K^{\text{input}}, i_K^{\text{output}}, i_{\text{test}}^{\text{input}}, i_{\text{pred}}),
\end{equation}
where $f^{\text{vision}}_{\text{critic}}$ assesses whether the test pair $(i_{\text{test}}^{\text{input}}, i_{\text{pred}})$ follows the same transformation pattern as the examples. The output $s_{\text{consistent}} \in \{\text{yes}, \text{no}\}$ indicates pattern consistency.

\noindent \textit{Step 3: Iterative Refinement.} If $s_{\text{consistent}} = \text{no}$, the model receives feedback about the inconsistency and performs another round of textual inference with error awareness:
\begin{equation}
    t_{\text{pred}} \leftarrow f^{\text{text}}_{\text{inf}}(r_{\text{pred}}, t_1^{\text{input}}, t_1^{\text{output}}, \ldots, t_{\text{test}}^{\text{input}}, \text{feedback}_{\text{prev}}),
\end{equation}
where $\text{feedback}_{\text{prev}}$ contains information about the previous attempt.
This process repeats until consistency is achieved or the iteration limit is reached (we use $N_{\max} = 3$).

\noindent \textbf{Key Advantages.} MSSC provides two critical benefits: (1) \textit{Fresh perspective}: switching to visual verification breaks the model's confirmation bias that occurs when checking its own textual reasoning, and (2) \textit{No external information needed}: unlike traditional self-correction that requires ground truth or external critics, MSSC uses the model's own multimodal capabilities.
As we show in \cref{sec:exp}, text-only self-correction often fails or even degrades performance, while MSSC achieves consistent improvements.
\label{sec:analysis}
\section{Experiments}
\label{sec:exp}
\subsection{Experimental setup}
\label{sec:setting}
\textbf{Models.} We evaluate both open-source and closed-source models.
For open-source models, we use Qwen3-VL-235B-A22B-Instruct~\cite{yang2025qwen3}.
For closed-source models, we use GPT-4o~\cite{hurst2024gpt}, Gemini-2.5-pro-thinking-8192~\cite{comanici2025gemini}, and o4-mini~\cite{openai2025o3o4mini}. 

\noindent \textbf{Benchmarks.} We evaluate on three ARC-AGI benchmarks.
\textbf{(1)} the official 400-task ARC-AGI \cite{chollet2019measure} evaluation set, \textbf{(2)} 100 randomly sampled tasks from Re-ARC \cite{hodel2024addressing} and \textbf{(3)} 100 randomly sampled tasks from BARC \cite{li2024combining}.
For Re-ARC and BARC, each sampled task contains four input-output pairs: three examples and one test instance.

\noindent \textbf{Implementation Details.}
We report Pass@1 accuracy across all experiments at a temperature 0.7.
Our prompts are provided in \cref{app:prompts} in the supplementary materials.

\subsection{Main Results}

\begin{table}[!t]
  \caption{\textbf{Visual-Language co-reasoning outperforms single-modality reasoning.} Both Vision-Language Synergy Reasoning (VLSR) and Modality-Switch Self-Correction (MSSC) improve over text-only baseline reasoning across models and benchmarks. The combination of both strategies yields the largest gains.}
  \vspace{-6pt}
  \centering
  \resizebox{0.49\textwidth}{!}{
  \normalsize
  \begin{tabular}{l|c|c|c}
    \toprule
    Models  & ARC-AGI & BARC-100 & Re-ARC\\
    \midrule
    GPT-4o & 8.25 & 28.0 & 10.0 \\
    \rowcolor{gray!6}\textit{+VLSR} & 13.5 & 32.0 & 13.0 \\
    \rowcolor{gray!6}\textit{+MSSC} & 12.0 & 30.0 & 14.0 \\
   \rowcolor{green!3} \textit{+both (Ours)} & 14.5 & 33.0 & 16.0 \\
    \midrule
    Gemini-2.5-Pro & 35.0 & 56.0 & 30.0 \\
    \rowcolor{gray!6}\textit{+VLSR} & 38.75 & 58.0 & 32.0 \\
    \rowcolor{gray!6}\textit{+MSSC} & 36.5 &  57.0 &  30.0 \\
   \rowcolor{green!3} \textit{+both (Ours)} & 42.25 & 60.0 & 33.0 \\
    \midrule
    o4-mini & 42.25 & 59.0 & 36.0 \\
    \rowcolor{gray!6}\textit{+VLSR} & 45.5 & 64.0 & 38.0 \\
    \rowcolor{gray!6}\textit{+MSSC} & 44.75 & 62.0 & 38.0 \\
   \rowcolor{green!3} \textit{+both (Ours)} & 46.75 & 65.0 & 39.0 \\
   \midrule
    Qwen3-VL-235B & 20.25 & 52.0 & 20.0 \\
    \rowcolor{gray!6}\textit{+VLSR} & 22.0 & 52.0 & 21.0  \\
    \rowcolor{gray!6}\textit{+MSSC} & 21.75 & 54.0 & 21.0 \\
   \rowcolor{green!3} \textit{+both (Ours)} & 22.25 & 54.0 & 23.0 \\
    \bottomrule
  \end{tabular}
  }
  \vspace{-6pt}
  \label{tab:main}
\end{table}

\noindent \textbf{Overall Performance.}
\cref{tab:main} shows that both VLSR and MSSC consistently improve performance across all models and benchmarks.
VLSR, which uses visual reasoning for rule summarization, improves baseline text-only reasoning by an average of 3.02\%.
MSSC, which uses visual verification for self-correction, provides an additional 1.82\% improvement on average.
Combining both strategies yields the largest gains, improving baseline performance by up to 6.25\% for GPT-4o and 7.25\% for Gemini-2.5-Pro on ARC-AGI.
These consistent improvements across diverse models demonstrate that visual information provides complementary benefits for both the rule summarization and the iterative refinement stages of abstract reasoning.

\noindent \textbf{Comparison with Training-Free Reasoning Methods.}
\begin{table}[!t]
  \caption{\textbf{Comparison with training-free inference-time methods.}
  All methods use o4-mini as the base model.
  Our VLSR+MSSC approach outperforms memory-augmented text-only strategies that retrieve past problem-solving experiences.
  }
  \vspace{-6pt}
  \centering
  \resizebox{0.49\textwidth}{!}{
  \normalsize
  \begin{tabular}{l|c|c|c}
    \toprule
    Method &ARC-AGI & ARC-AGI-100 & Re-ARC \\
    \midrule
     Direct Reason & 40.5 & 41.0 &36.0 \\
     Cheatsheet & 38.5 & 41.0 & 34.0 \\
     ArcMemo-PS & 45.25 & 45.0 & 39.0 \\
     \textbf{Ours} & 46.75 & 46.0 & 39.0\\
    \bottomrule
  \end{tabular}
  }
  \label{tab:comparing}
  \vspace{-6pt}
\end{table}
We compare against two recent training-free strategies that enhance ARC-AGI reasoning at inference time:
1) \textbf{Dynamic Cheatsheet} \cite{suzgun2025dynamic} stores strategies and findings from past problem-solving processes as memory and includes them in prompts for new problems.
2) \textbf{ArcMemo-PS} \cite{ho2025arcmemo} builds concept-level external memory through program synthesis and extracts reusable modular abstract concepts, and selectively retrieves relevant concepts during reasoning.
Both methods are text-centric.

Using o4-mini as the base model, we evaluate all methods on three test sets: the full ARC-AGI-400, the ARC-AGI-100 subset from ArcMemo \cite{ho2025arcmemo}, and the Re-ARC test set we divided.
As shown in \cref{tab:comparing}, our method achieves the highest accuracy across all three benchmarks, outperforming the strongest baseline ArcMemo-PS by 1.5\% on ARC-AGI.
Notably, while ArcMemo-PS and Cheatsheet use retrieved past experiences for text-only reasoning, they cannot access the global 2D structure and spatial pattern information that visual representations provide.
Our results suggest that visual information offers complementary benefits that text-based memory retrieval alone cannot capture.

\begin{figure}[t]
  \centering
  \includegraphics[width=\linewidth]{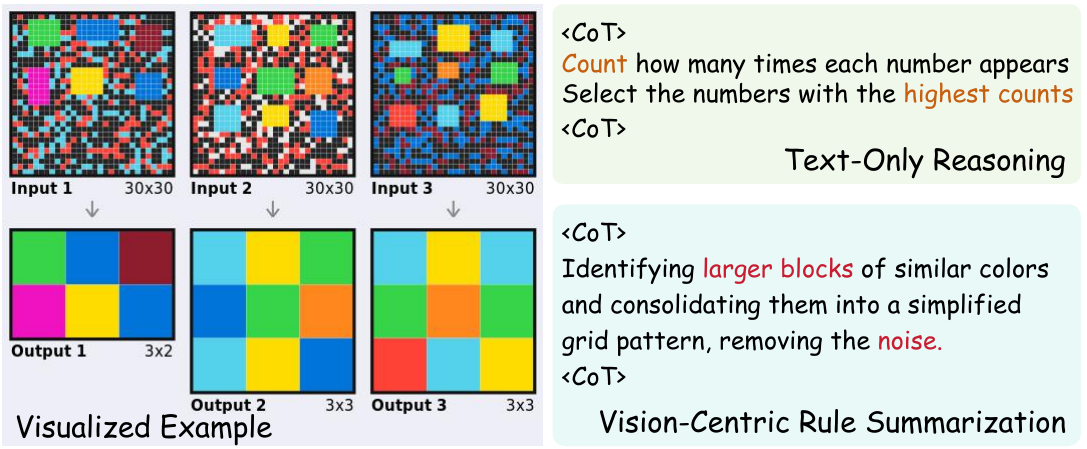}
  \vspace{-10pt}
   \caption{\textbf{Qualitative comparison of text-only vs. vision-language synergy reasoning on GPT-4o.}
   Text-only reasoning processes elements without spatial context, leading to an incorrect rule.
   Vision-language synergy reasoning uses global 2D perception in the rule-summarization phase to identify the correct spatial pattern (``retain large connected color blocks'').}
   \vspace{-6pt}
   \label{fig:case}
\end{figure}

\subsection{Analysis}
\begin{table}[!t]
  \caption{
  \textbf{Iterative self-correction comparison.}
  We apply Text-Only Self-Correction (TOSC) and Modality-Switch Self-Correction (MSSC) for up to three rounds (R1-R3) without external feedback. MSSC achieves consistent iterative improvement while TOSC stagnates or degrades.
  }
  \vspace{-6pt}
  \centering
  \resizebox{0.49\textwidth}{!}{
  \normalsize
  \begin{tabular}{l|c|c|c|c|c|c|c}
    \toprule
    \multirow{2}{*}{Models} & \multirow{2}{*}{Base} & \multicolumn{3}{c|}{TOSC} & \multicolumn{3}{c}{MSSC} \\
    \cmidrule(lr){3-5} \cmidrule(lr){6-8} 
     &  & R1 & R2 & R3 & R1 & R2 & R3 \\
    \midrule
     GPT-4o & 8.25 & 8.25 & 8.0 & 8.75 & 10.25 & 11.5 & 12.0 \\
     Gemini & 35.0 & 34.25 & 36.0 & 35.75 & 35.75 & 36.25 & 36.5 \\
     o4-mini & 42.25 & 42.5 & 42.0 & 43.25 & 43.5 & 44.25 & 44.75 \\
    \bottomrule
  \end{tabular}
  }
  \vspace{-6pt}
  \label{tab:mssc}
\end{table}

\noindent \textbf{Modality-Switch vs. Text-Only Self-Correction}
\cref{tab:mssc} compares traditional text-only self-correction (TOSC) with our Modality-Switch Self-Correction (MSSC) across three iterative rounds.
TOSC shows minimal improvement and even degrades performance in some iterations.
For GPT-4o, TOSC improves by only 0.5 points (8.25$\rightarrow$8.75) across three rounds, with Round 2 degrading to 8.0.
In contrast, MSSC achieves consistent monotonic gains at each round. For GPT-4o, MSSC improves from 8.25 to 10.25 (+2.0) in Round 1, then to 11.5 (+1.25) in Round 2, and finally to 12.0 (+0.5) in Round 3.
Similar trends hold for Gemini (+1.5 total) and o4-mini (+2.5 total).

We attribute this difference to modality switching: verifying textual outputs with visual representations provides a \textit{fresh perspective} that helps models identify spatial inconsistencies imperceptible when reasoning solely in text.
When TOSC uses the same textual modality for both generation and verification, the model exhibits confirmation bias and cannot effectively spot its own errors.
By contrast, MSSC's visual verification stage detects pattern violations (e.g., missing symmetry, incorrect spatial relationships) that the textual reasoning stage overlooked, enabling genuine iterative improvement without external feedback.

\noindent \textbf{Qualitative Analysis.}
\cref{fig:case} presents how visual reasoning corrects systematic errors made by text-only reasoning on GPT-4o.
When presented with textual matrix representations, GPT-4o incorrectly summarizes the transformation rule as ``select the number with the highest count''.
This frequency-based heuristic fails because the task actually requires identifying large connected spatial structures.
With visualized matrices, the model correctly identifies the rule as ``retain large connected color blocks'' by using global 2D perception to recognize spatial contiguity patterns.
Results show that visual representation enables holistic pattern recognition.
Additional examples are provided in \cref{app:qualitative} in supplementary materials.

\subsection{Extension to Model Fine-tuning}
\begin{table}[!t]
  \caption{\textbf{Experiments on fine-tuning Results.}
  Vision-Language Synergy Fine-tuning outperforms both text-only fine-tuning (using the same training tasks) and several closed-source and open-source baseline models with much larger parameter sizes.
  }
  \vspace{-6pt}
  \centering
  \resizebox{0.49\textwidth}{!}{
  \normalsize
  \begin{tabular}{l|c|c|c}
    \toprule
    Models  & ARC-AGI & BARC-100 & Re-ARC\\
    \midrule
    \rowcolor{blue!5}
    \multicolumn{4}{c}{\textit{Text-only Reasoning Baseline}} \\
    \midrule
    GPT-4o & 8.25 & 28.0 & 10.0 \\
    Gemini-2.5-Pro & 35.75 & 56.0 & 30.0 \\
    o4-mini & 42.25 & 59.0 & 36.0 \\
    Qwen3-235B-A22B & 20.25 & 52.0 & 20.0\\  
    \rowcolor{first1}
    Qwen3-8B & 3.25  & 13.0 & 2.0 \\
    \rowcolor{first2} 
   \textit{+Text-only FT} & 9.75  & 37.0  & 7.0 \\
   \rowcolor{first3} \textit{Improvement} & \textcolor[RGB]{28, 96, 67}{+6.5} &\textcolor[RGB]{28, 96, 67}{+24.0} & \textcolor[RGB]{28, 96, 67}{+5.0} \\
   \midrule
   \rowcolor{blue!5}
   \multicolumn{4}{c}
   {\textit{Vision-Language Synergy Reasoning}} \\
   \midrule
   \rowcolor{first1}
   Qwen3-VL-8B + Qwen3-8B& 3.5  & 13.0 & 3.0 \\
    \rowcolor{first2}
    \textit{+VL Synergy FT} & 13.25&43.0 & 9.0\\
    \rowcolor{first3}
    \textit{Improvement} & \textcolor[RGB]{28, 96, 67}{+9.75} & \textcolor[RGB]{28, 96, 67}{+30.0} & \textcolor[RGB]{28, 96, 67}{+6.0} \\

    \bottomrule
  \end{tabular}
  }
  \vspace{-6pt}
  \label{tab:training}
\end{table}

\textbf{Background:}
Open-source models typically underperform on ARC-AGI due to limited capabilities on abstract reasoning tasks.
Prior work \cite{hodel2024addressing, li2024combining, akyurek2024surprising, franzen2024llm} constructs large-scale synthetic training data to fine-tune open-source models, but treats ARC-AGI as a pure text task without using visual information.
We investigate whether vision-language synergy during fine-tuning can improve open-source model performance beyond text-only fine-tuning.

\noindent 
\textbf{Method.}
We apply the same VLSR task decomposition from \cref{sec:vlsr} during training: a vision-language model (Qwen3-VL-8B-Instruct) for visual rule summarization and a text-only model (Qwen3-8B) for textual rule application.
This decouples the two subtasks, allowing for specialized training of each component.

\noindent
\textbf{Training Setup.}
Training data comes from ARC-Heavy-200k \cite{li2024combining}, which provides synthetic ARC-AGI tasks with ground-truth rules.
We use approximately 200k training tasks (excluding a held-out 100-task test set), with each task split into three example pairs and one test sample.
ARC-Heavy-200k provides explicit rules for each task, enabling us to train the rule summarization and rule application modules separately.
As a text-only baseline, we also fine-tune Qwen3-8B on the same 200k tasks using textual matrix representations from ARC-Heavy-200k.

\noindent
\textbf{Results.}
As shown in \cref{tab:training}, VL synergy fine-tuning achieves 13.25\% on ARC-AGI, outperforming text-centric fine-tuning (9.75\%) by 3.5\% and surpassing the closed-source baseline GPT-4o (8.25\%) by 5.0\%.
Compared to the Qwen3-8B baseline before fine-tuning, text-only fine-tuning improves performance by 6.5\% while VL synergy fine-tuning improves by 9.75\%, which demonstrates the advantage of incorporating visual information during training.
Similar trends hold on BARC-100 and Re-ARC.
We attribute the advantage of VL synergy fine-tuning to two factors: (1) task decomposition reduces training complexity by separating rule extraction from rule application, and (2) visual information provides global 2D structural cues (e.g., spatial contiguity, symmetry patterns) that are difficult to learn from sequential textual representations alone.
\section{Conclusion}

This paper incorporates visual information into abstract reasoning tasks, which have traditionally been treated as purely textual.
By proposing the Vision-Language Synergy Reasoning (VLSR) framework and the Modality-Switch Self-Correction (MSSC) mechanism, the 2D intuitiveness inherent in the visual modality and the precision of element-wise representation offered by the textual modality can be effectively integrated.
Leveraging the distinct advantages of both modalities, our method achieves an average performance improvement of 4.3\% compared to text-only reasoning approaches across multiple abstract reasoning tasks and different base models.
Furthermore, it outperforms other text-centric training-free strategies, thereby demonstrating the unique value of incorporating visual information.
Additionally, this inference strategy can be extended to model fine-tuning scenarios.
The vision-language synergy fine-tuning strategy can yield a 3.5\% improvement over text-only fine-tuning methods and outperform several open-source and closed-source models with much larger parameter sizes.

{
    \small
    \bibliographystyle{ieeenat_fullname}
    \bibliography{main}
}

\appendix
\clearpage
 \setcounter{page}{1}
 \maketitlesupplementary

\section{Prompts}\label{app:prompts}
\textbf{Prompt for Text-only Reasoning in ARC-AGI:}

\noindent
I will provide you with several input and output matrices. You need to find the matrix-changing rule from it and  apply it to the new input. Put the output matrix within \textbackslash\textbackslash boxed\{\}.\\

\noindent
Example Input 1: [[0, 1, 0, ...], [0, 1, 1, ...], ...]

\noindent
Example Output 1: [[1, 1, 0, ...], [1, 1, 1, ...], ...]

\noindent
Example Input 2: [[0, 2, 4, ...], [0, 2, 2, ...], ...]

\noindent
Example Output 2: [[2, 2, 4, ...], [2, 2, 2, ...], ...]

\noindent
...

\noindent
Example Input n: [[0, 4, 0, ...], [0, 6, 1, ...], ...]

\noindent
Example Output n: [[4, 4, 0, ...], [6, 6, 1, ...], ...]

\noindent
New Input: [[0, 2, 0, ...], [0, 5, 3, ...], ...]
\\

\noindent
\textbf{Prompt for Vision-centric Rule Summarization:}

\noindent
I will now provide you with several input and output images about 2D grids. You need to summarize the grid-changing rule from it. Output the rule you learned within \textbackslash\textbackslash boxed\{\}.\\

\noindent
Example Input 1: \textless Input Image 1\textgreater

\noindent
Example Output 1: \textless Output Image 1\textgreater

\noindent
Example Input 2: \textless Input Image 2\textgreater

\noindent
Example Output 2: \textless Output Image 2\textgreater

\noindent
...

\noindent
Example Input n: \textless Input Image n\textgreater

\noindent
Example Output n: \textless Output Image n\textgreater
\\

\noindent
\textbf{Prompt for Text-centric Rule Application:}

\noindent
I will provide you with several input and output matrices. You need to find the matrix-changing rule from it and  apply it to the new input. Put the output matrix within \textbackslash\textbackslash boxed\{\}.
\\

\noindent
Here is a possible rule for your reference. \textit{Rule: The rule involves removing the colored cross ...} Note that the rule is described in color and each color represents a value in the matrix: [0:black; 1:blue; 2:red; 3:green; 4:yellow; 5:grey; 6:pink; 7:orange; 8:light blue; 9:brown].
You need to first check the correctness of the rule based on the examples. If the rule is correct, apply it to the new input. Otherwise, summarize a new rule and apply it to the new input.\\

\noindent
Example Input 1: [[0, 1, 0, ...], [0, 1, 1, ...], ...]

\noindent
Example Output 1: [[1, 1, 0, ...], [1, 1, 1, ...], ...]

\noindent
Example Input 2: [[0, 2, 4, ...], [0, 2, 2, ...], ...]

\noindent
Example Output 2: [[2, 2, 4, ...], [2, 2, 2, ...], ...]

\noindent
...

\noindent
Example Input n: [[0, 4, 0, ...], [0, 6, 1, ...], ...]

\noindent
Example Output n: [[4, 4, 0, ...], [6, 6, 1, ...], ...]

\noindent
New Input: [[0, 2, 0, ...], [0, 5, 3, ...], ...]
\\

\noindent
\textbf{Prompt for Vision-centric Consistency Verification:}

\noindent
I will now provide you with several input and output example images, which follows a specific changing rule. Then, I will give you another input and output pair, determine whether the new pair also follows the same changing rule. Add your final judgment at the end of your replay: \textbackslash\textbackslash boxed\{True\} or \textbackslash\textbackslash boxed\{False\}.\\

\noindent
Example Input 1: \textless Input Image 1\textgreater

\noindent
Example Output 1: \textless Output Image 1\textgreater

\noindent
Example Input 2: \textless Input Image 2\textgreater

\noindent
Example Output 2: \textless Output Image 2\textgreater

\noindent
...

\noindent
Example Input n: \textless Input Image n\textgreater

\noindent
Example Output n: \textless Output Image n\textgreater

\noindent
New Input: \textless New Input Image\textgreater

\noindent
New Output: \textless Output Image Pred\textgreater

\section{Matrix-to-Image Visualization}\label{app:visualization}

We visualize the input-output matrices into color-coded 2D grids to provide 2D spatial information and global view. The detailed visualization process is listed as follows.

First, each value is mapped to a distinct color. The color map is:
[
    0:  black;
    1:  blue;
    2:  red;
    3:  green;
    4:  yellow;
    5:  grey;
    6:  pink;
    7:  orange;
    8:  light blue;
    9:  brown
]

Furthermore, between the two elements (small colored squares), we add white dividing lines to more clearly indicate the specific number and the structure of the elements contained within a block.

\section{Qualitative Examples}\label{app:qualitative}
\begin{figure*}[h]
  \centering
  \includegraphics[width=\linewidth]{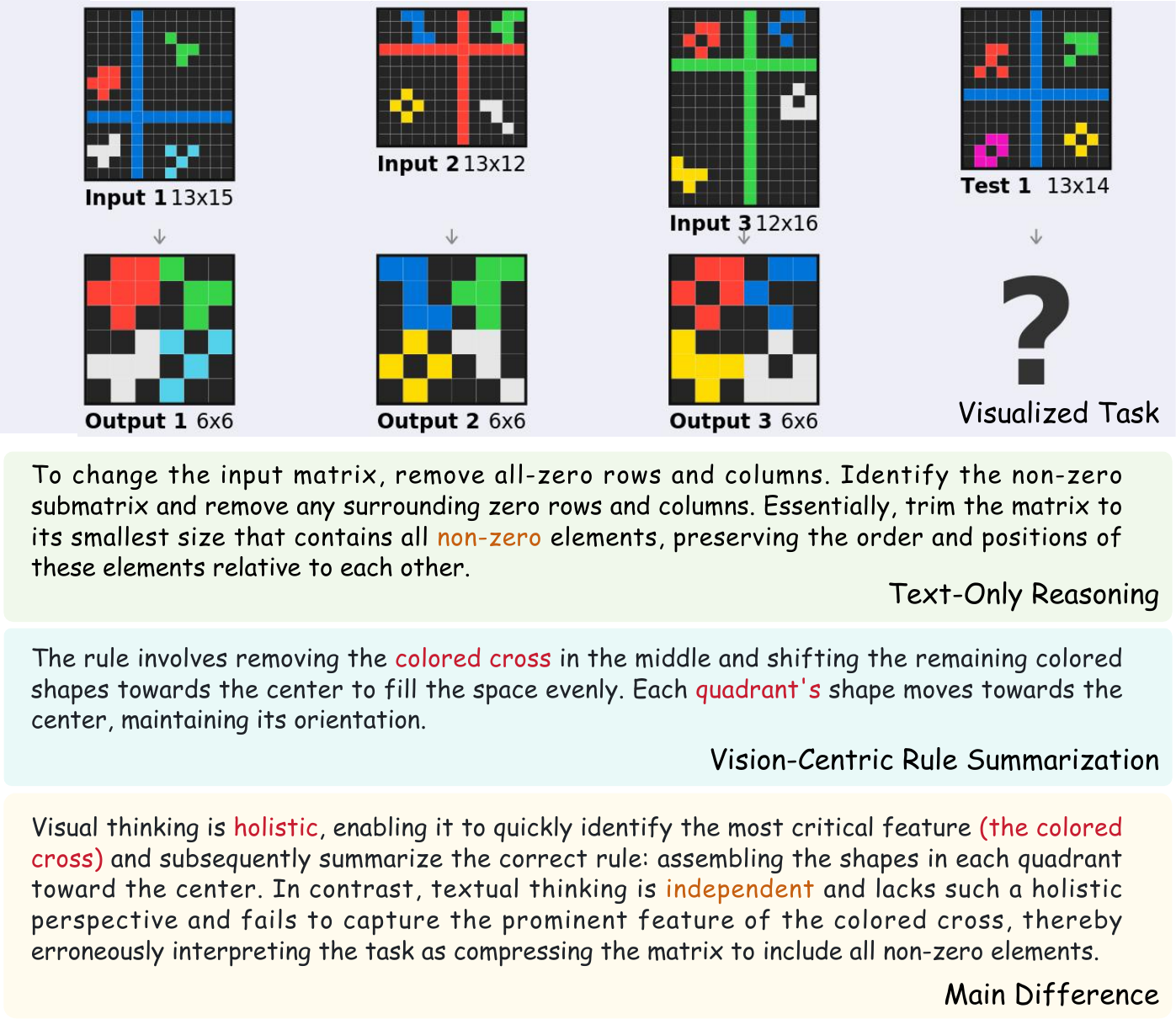}
   \caption{Visual reasoning possesses a global perspective, enabling it to better capture the most critical feature (the colored cross) in the entire image and subsequently summarize the correct rule the underlying rules. Base model is both gpt-4o.}
  \label{fig:eg1}
\end{figure*}

\begin{figure*}[h]
  \centering
  \includegraphics[width=\linewidth]{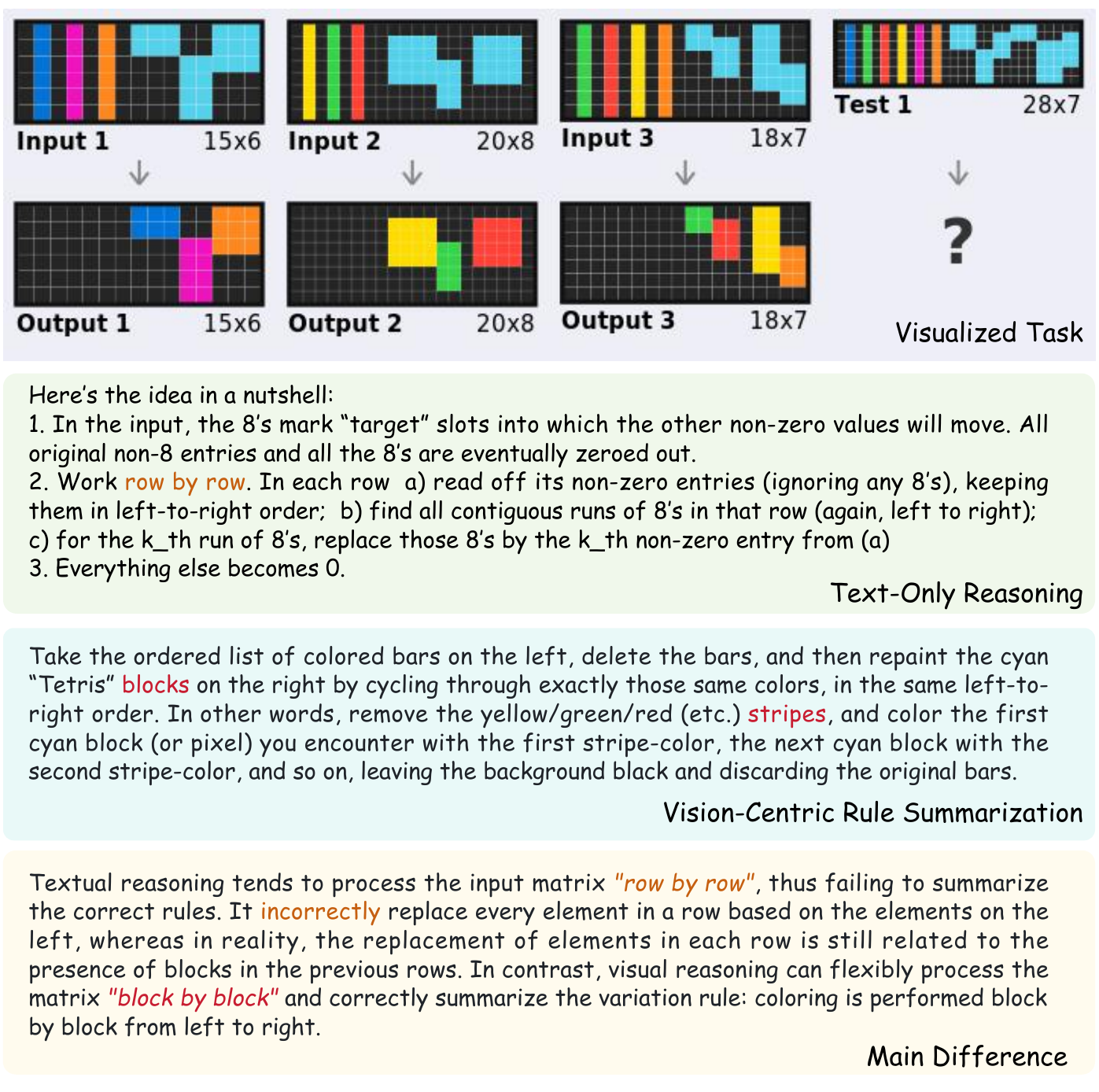}
   \caption{Visual reasoning possesses 2D information and can flexibly summarize rules in a ``block-by-block'' manner, whereas text reasoning tends to adopt a ``row-by-row'' processing approach, thus failing to derive the correct rules. Base model is both o4-mini.}
  \label{fig:eg2}
\end{figure*}

\begin{figure*}[h]
  \centering
  \includegraphics[width=\linewidth]{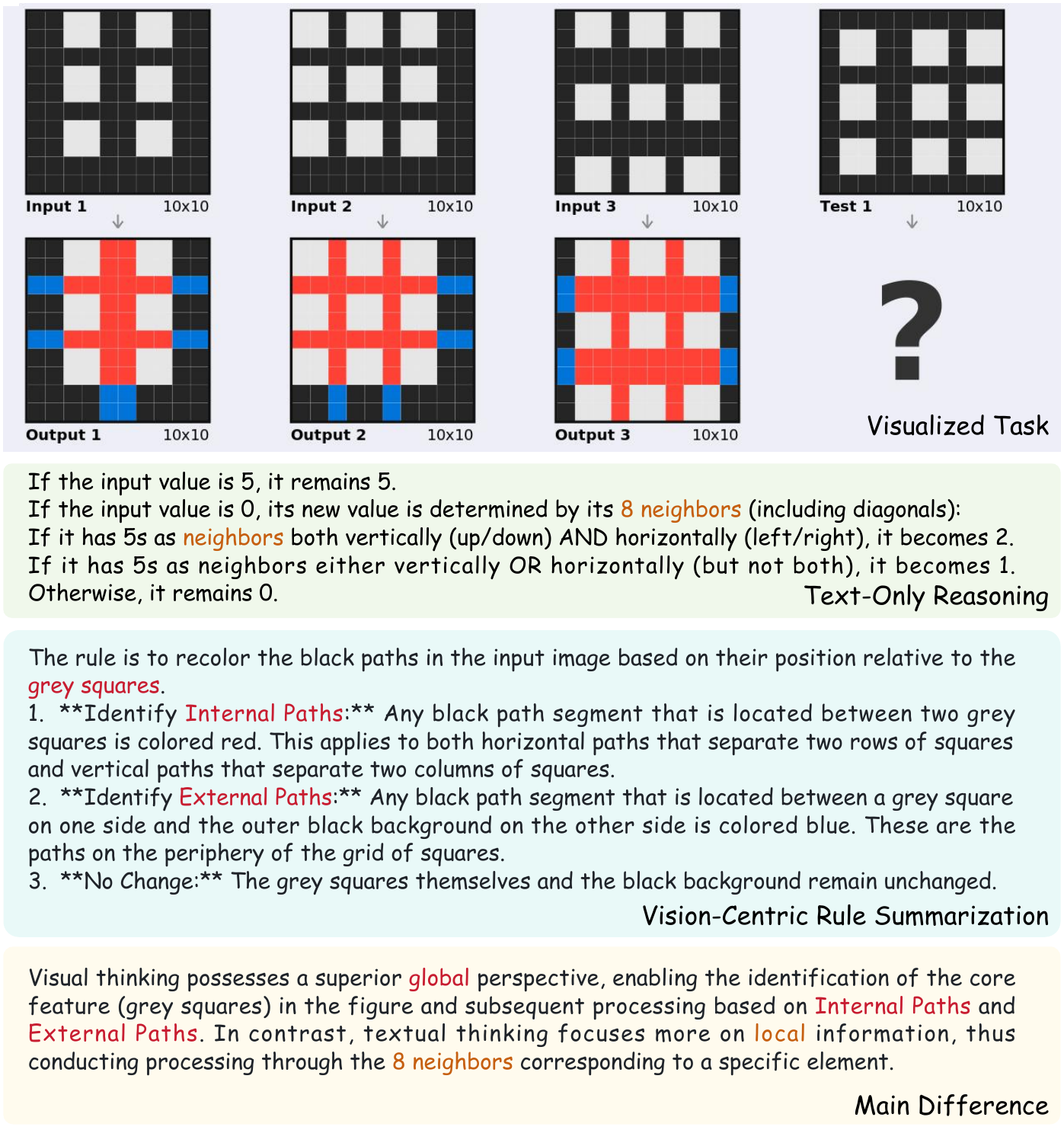}
   \caption{Visual thinking tends to adopt a global perspective and thus processes information based on internal and external paths; in contrast, textual thinking focuses more on local information and processes individual elements with reference to their 8-neighbor context. Base model is both Gemini-2.5-Pro-thinking-8192.}
  \label{fig:eg3}
\end{figure*}

\begin{figure*}[h]
  \centering
  \includegraphics[width=\linewidth]{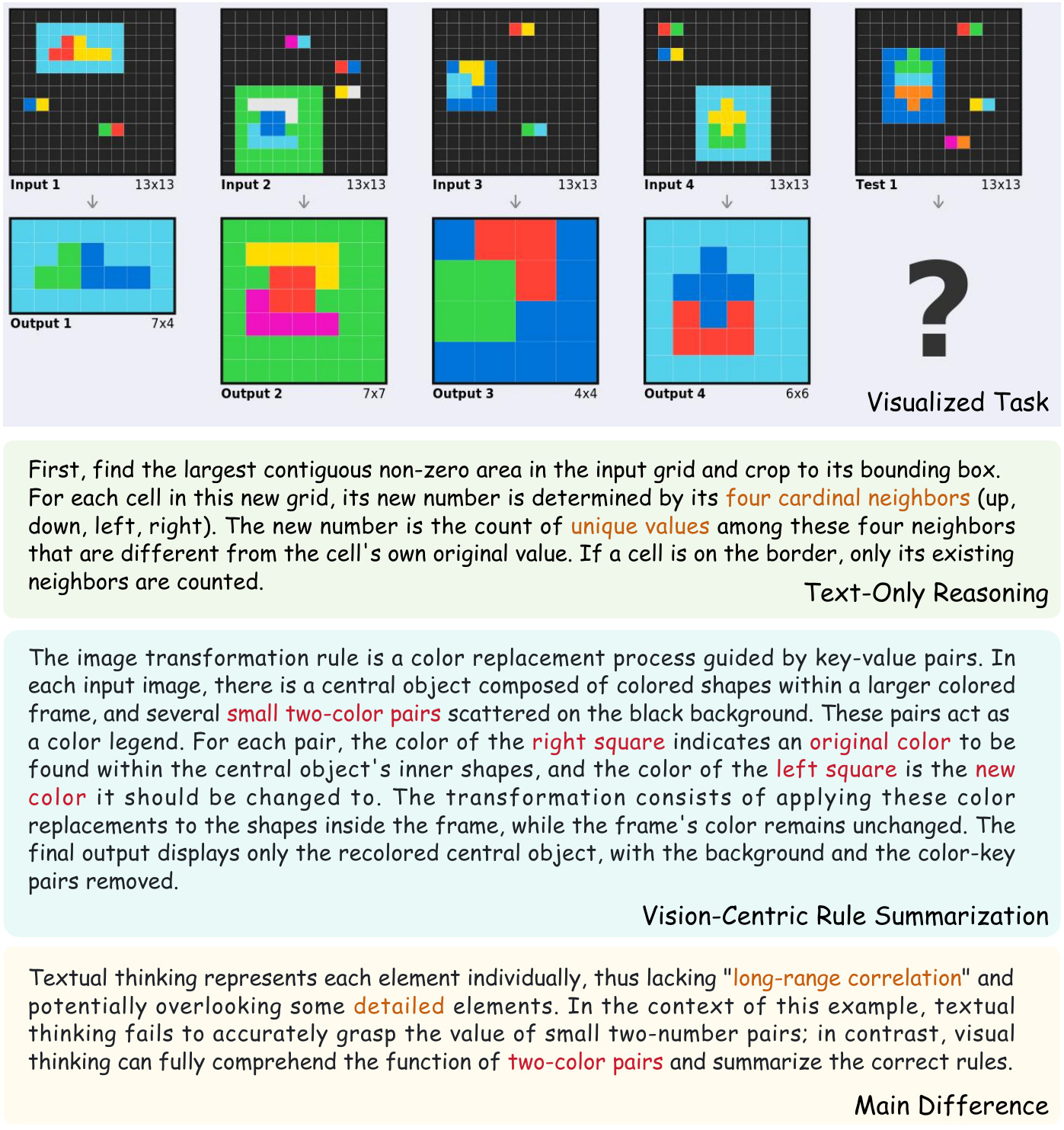}
   \caption{Visual thinking possesses superior long-range correlation capabilities and can better capture detailed features (the 2-color pairs and the re-color strategy). Base model is both Gemini-2.5-Pro-thinking-8192.}
  \label{fig:eg4}
\end{figure*}

We conduct an in-depth analysis of the specific outputs of different models (GPT-4o, Gemini-2.5-Pro-thinking-8192, o4-mini) when employing visual thinking versus textual thinking in the ARC-AGI task. Visual thinking demonstrates numerous unique advantages, such as the integration of 2D structural information(\cref{fig:eg2}), a global perspective(\cref{fig:eg1}, \cref{fig:eg3}), and long-range perception capabilities(\cref{fig:eg4}). Therefore, using visual information during the global rule summarization phase enables more effective extraction of the underlying rules.

\end{document}